\documentclass{article}
\PassOptionsToPackage{table}{xcolor}
\PassOptionsToPackage{authoryear,round}{natbib}
\usepackage[preprint]{neurips_2026}
\makeatletter
\renewcommand{\@noticestring}{Preprint.}
\makeatother
\ifdefined\pdfminorversion
\fi

\usepackage[utf8]{inputenc}
\usepackage[T1]{fontenc}
\usepackage{amsmath,amssymb}
\usepackage{graphicx,caption,wrapfig}
\usepackage{booktabs,array,tabularx,multirow,siunitx}
\usepackage{microtype,xcolor,pgfplots}
\pgfplotsset{compat=1.18}
\usepackage[hidelinks]{hyperref}
\setcitestyle{authoryear,round,citesep={;},aysep={,},yysep={;}}
\hypersetup{
  pdftitle={Towards Long-horizon Embodied Agents with Tool-Aligned Vision-Language-Action Models},
  pdfauthor={Zixing Lei, Changxing Liu, Yichen Xiong, Minhao Xiong, Yuanzhuo Ding, Zhipeng Zhang, Weixin Li, Siheng Chen}
}

\definecolor{TableHeader}{RGB}{235,240,246}
\definecolor{TableHighlight}{RGB}{243,247,251}
\definecolor{TableRule}{RGB}{88,101,116}
\definecolor{TableGain}{RGB}{200,40,40}
\newcommand{\modernTableSetup}{%
  \setlength{\aboverulesep}{0pt}%
  \setlength{\belowrulesep}{0pt}%
  \setlength{\extrarowheight}{0pt}%
  \arrayrulecolor{TableRule}%
  \renewcommand{\arraystretch}{1.08}%
  \renewcommand{\tabularxcolumn}[1]{m{##1}}%
}
\newcolumntype{L}{>{\raggedright\arraybackslash}X}
\newcommand{\gain}[1]{\textcolor{TableGain}{($\uparrow$#1)}}

\title{Towards Long-horizon Embodied Agents with Tool-Aligned Vision-Language-Action Models}

\author{%
  Zixing Lei\textsuperscript{1,2}\thanks{Equal contribution} \quad
  Changxing Liu\textsuperscript{1}\footnotemark[1] \quad 
  Yichen Xiong\textsuperscript{1} \quad 
  Minhao Xiong\textsuperscript{1} \quad 
  Yuanzhuo Ding\textsuperscript{1} \\ 
\textbf{Zhipeng Zhang}\textsuperscript{1} \quad 
  \textbf{Weixin Li}\textsuperscript{2,3} \quad 
  \textbf{Siheng Chen}\textsuperscript{1}\thanks{Corresponding author}
\\
\textsuperscript{1}Shanghai Jiao Tong University  \quad
\textsuperscript{2}Zhongguancun Academy \quad 
\\
\textsuperscript{3}Beihang University \quad 
\\
{\tt\small chezacarss@sjtu.edu.cn \quad cx-liu@sjtu.edu.cn \quad sihengc@sjtu.edu.cn}
}

\begin{document}
\raggedbottom

\maketitle
\begin{abstract}
  Vision-language-action (VLA) models are effective robot action executors, but they remain limited on long-horizon tasks due to the dual burden of extended closed-loop planning and diverse physical operations. We therefore propose \textbf{VLAs-as-Tools}, a strategy that distributes this burden across a high-level vision language model (VLM) agent for temporal reasoning and a family of specialized VLA tools for diverse local physical operations. The VLM handles scene analysis, global planning, and recovery, while each VLA tool executes a bounded subtask. To tightly couple agent planning with VLA tool execution in long-horizon tasks, we introduce a \emph{VLA tool-family interface} that exposes explicit tool selection and in-execution progress feedback, enabling efficient event-triggered agent replanning without continuous agent polling. To obtain diverse specialized VLA tools that faithfully follow agent invocations, we further propose \emph{Tool-Aligned Post-Training} (TAPT), which constructs invocation-aligned training units for instruction following and adopts tool-family residual adapters for efficient tool specialization. Across five simulated benchmarks, VLAs-as-Tools improves the success rate of $\pi_{0.5}$ by 8.9 percentage points on average and enhances invocation fidelity by 15.0 points in Non-biased Rate. On three real-world Franka tasks, it improves success rate from 16.7\% to 58.3\% under matched data and compute. Code will be released.
\end{abstract}

\vspace{-2mm}
\section{Introduction}
\vspace{-2mm}

Recent end-to-end vision-language-action (VLA) models, including OpenVLA-OFT and GR00T N1.6, have advanced general-purpose robot control and cross-embodiment generalization~\citep{openvlaoft, grootn16}. Yet they are typically deployed as single embodied controllers, requiring one policy to interpret goals, perceive scenes, predict actions, and coordinate multi-step execution. This setting is difficult for current VLAs: long-horizon tasks require decomposition, progress tracking, failure recovery, and skill composition, while embodied data remains limited and model scale still constrains planning and understanding. As a result, VLAs provide powerful language-conditioned robot-control capabilities, but remain limited as standalone long-horizon agents.

There are two routes toward long-horizon embodied tasks. One direction keeps planning inside end-to-end VLA policies, where models such as $\pi_{0.5}$ introduce semantic subtask prediction and task-level planning tokens within the VLA backbone~\citep{pi05}. Another direction, represented by SayCan and Code as Policies, adopts an agentic formulation in which language models select skills, compose programs, or invoke robot APIs~\citep{saycan,codepolicies}. This follows the successful paradigm of digital agents, where high-capacity LLMs handle goal decomposition, state tracking, replanning, and error recovery, while tools provide bounded, observable, and reliable execution interfaces~\citep{react, toolformer, sweagent}. However, existing embodied tools are often manually specified and task-specific, giving them weaker physical execution capabilities than modern VLAs. This leaves embodied AI with a complementary gap: end-to-end VLAs provide stronger physical execution but weaker long-horizon reasoning, whereas existing embodied agents provide stronger planning but rely on weaker physical tools.

\begin{figure*}[t]
    \centering
    \includegraphics[width=\linewidth]{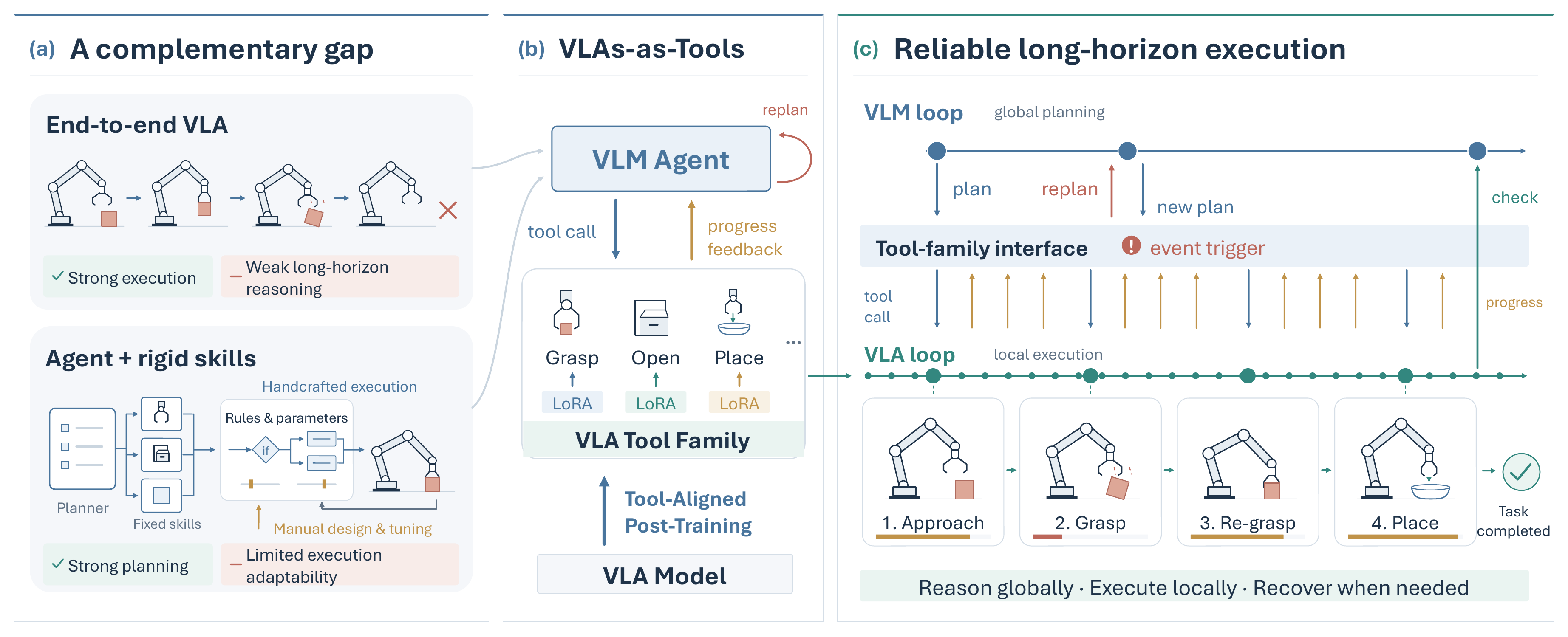}
    \caption{\emph{VLAs-as-Tools} closes the complementary gap between end-to-end VLAs, which execute well but reason poorly over long horizons, and agent systems, which plan well but rely on weak rigid skills. It couples a planning VLM with TAPT-specialized VLA tools through explicit tool calls, progress feedback, and event-triggered replanning for reliable long-horizon execution.}
    \label{fig:overview}
    \vspace{-5mm}
\end{figure*}

As illustrated in Figure~\ref{fig:overview}, we propose \emph{VLAs-as-Tools}, a strategy for solving long-horizon embodied tasks by enabling a planning-capable VLM agent to invoke bounded VLA executions as physical tools. This strategy addresses the limitations of individual VLAs by distributing the burden across time length and task breadth: the VLM extends execution over long horizons through planning, state tracking, and recovery, while multiple specialized VLA tools cover diverse bounded subtasks. Each VLA therefore operates within a focused execution scope, where it can be efficiently adapted from task-specific demonstrations and reliably map visual-language inputs to continuous robot actions. However, standalone VLAs are not readily tool-usable, since their execution can be biased by scene priors, demonstration regularities, or visual context rather than the invoked instruction~\citep{RLG,LibroCF}, and task-specific adaptation can degrade pretrained semantic understanding and generalization~\citep{vlm2vla,lifelongrft}. Thus, realizing \emph{VLAs-as-Tools} requires addressing two core challenges: how to make VLAs callable and reliable physical tools, and how to integrate them effectively into an embodied agentic system.

We address these two challenges with two complementary components. To make VLAs tool-usable, we introduce \emph{Tool-Aligned Post-Training} (TAPT), which pairs bounded subtasks with precise language instructions so that each VLA tool learns a clear correspondence between the agent's invocation and the intended physical behavior, and adopts tool-family residual adapters to parameter-efficiently create many specialized VLA tools on a shared backbone while mitigating degradation of pretrained semantic generalization. To integrate these tools into the agentic system, we design a \emph{VLA tool-family interface} with two typed message sets: agent-to-tool invocations that specify bounded subtasks and route them to the appropriate VLA tool, and tool-to-agent feedback that returns in-execution state signals. The former grounds planned subtask sequences in the correct physical executor, while the latter exposes timely intervention points, enabling low-frequency but responsive replanning.

We evaluate three VLA backbones---OpenVLA, OpenVLA-OFT, and $\pi_{0.5}$---on six benchmarks: LIBERO-Long, RoboTwin-SA, CALVIN, RoboMemArena, RoboLab-120, and LIBERO-IF, plus three real-world Franka tasks. Across the five simulated task-success benchmarks with directly matched $\pi_{0.5}$ baselines, \emph{VLAs-as-Tools} improves average success by 8.9 points. It also improves invocation fidelity by 15.0 points in Non-biased Rate and raises real-world macro-average success from 16.7\% to 58.3\%.

\vspace{-2mm}
\section{Related Work}
\vspace{-2mm}

\paragraph{VLA Tool Interfaces for Agentic Embodied Control.}
Recent progress in embodied foundation models has been driven by Vision-Language-Action (VLA) policies, which map visual observations and language instructions to robot actions with large pretrained vision-language backbones~\citep{rt2,octo,openvla,pi0,pi05}. While these models achieve strong full-task performance, studies show that VLAs do not always ground their behavior in the language command, instead following visual priors and training-set regularities~\citep{bcz,whatmatters,cast,liberoplus}; counterfactual evaluations further show visual cues overriding the specified instruction~\citep{LibroCF}. This makes standalone VLAs unreliable as agentic execution interfaces, where the planner needs each invocation to have stable, instruction-faithful semantics that can be monitored and composed. We address this by aligning VLA post-training with the agent's invocation interface, using subtask-centric data, invocation-aligned training, and tool-family residual adapters to bind each command and tool-family label to a bounded execution behavior.

\paragraph{Embodied Agentic Systems}
Recent embodied AI advances have increasingly focused on agentic systems, particularly in language-conditioned and long-horizon tasks~\citep{embodiedagenticai,largemodelembodiedsurvey}. Hierarchical embodied systems commonly address such tasks by decomposing high-level goals into task-level decisions and executable low-level skills~\citep{saycan,innermono,rth,hirobot}, but they often rely on hand-engineered skills, affordance models, latent options
or jointly trained executors. Recent systems further scale this paradigm with learned skill libraries and agentic execution loops: AtomicVLA builds a planning-and-execution framework around atomic skill abstractions and skill-guided expert specialization~\citep{atomicvla}, while RoboClaw uses a VLM-driven controller to orchestrate learned policy primitives for long-horizon tasks and autonomous data collection~\citep{roboclaw}. These works validate skill decomposition and policy composition but their low-level skills are often system-internal primitives rather than explicit agent-visible tool invocations. In contrast, we expose the VLA as a reusable embodied tool whose agent-visible invocations pair bounded language instructions with tool-family labels, enabling more inspectable and composable execution.

\vspace{-2mm}
\section{Formalizing VLAs as Callable Tools}
\vspace{-2mm}
\label{sec:vla-as-tool-formalization}

This section formalizes the \emph{VLAs-as-tools} strategy: a VLA is not used as a monolithic long-horizon policy, but as a family of bounded, callable executors inside an embodied agentic control loop. The connection between the high-level agent and the VLA tools is an interface $ \mathcal{I}=(\mathcal{C},\mathcal{R}),$
where $\mathcal{C}$ is the set of agent-to-tool invocation messages and $\mathcal{R}$ is the set of tool-to-agent feedback messages. Given a user goal $q$, a high-level agent $\Pi_\phi$ tracks an agent-side state $s_k$ over observations, past calls, and returned feedback. At step $k$, the agent sends an invocation message via the interface,
\begin{equation}
    c_k = \Pi_\phi(q,s_k),
    \qquad
    c_k=(g_k,z_k)\in\mathcal{C}=\mathcal{G}\times\mathcal{Z},
\end{equation}
where $g_k$ is a discrete tool-family label, such as \textit{reach}, \textit{grasp}, or \textit{pull}, and $z_k$ is a scene-grounded subtask instruction. The tool-family label $g_k$ selects a member of the VLA tool family,$\mathcal{T}=\{T_g\}_{g\in\mathcal{G}},$
where each $T_g$ is a callable VLA tool for one tool family. The instruction $z_k$ grounds it in the current scene by specifying the object, relation, and desired local effect.

The selected tool $T_{g_k}$ executes the call over a bounded low-level horizon. Given robot observations $o_t$, it produces actions and a bounded trajectory
\begin{equation}
    a_t \sim T_{g_k}(\cdot \mid o_t,z_k,h_t),
    \qquad
    \tau_k=(o_{t_k:t_k+H_k},a_{t_k:t_k+H_k-1}),
\end{equation}
where $h_t$ denotes an optional short execution history and $H_k$ is the call horizon. After or during execution, the selected tool returns feedback $r_k$ such as progress or completion information. The agent state is then updated as
\begin{equation}
    s_{k+1}=U(s_k,c_k,\tau_k,r_k),
    \qquad
    r_k \in \mathcal{R},
\end{equation}
and the loop repeats until the task terminates. Thus, the interface defines both directions of communication: $\mathcal{C}$ tells the selected tool what to execute, and $\mathcal{R}$ tells the agent what happened during execution. The invocation message set $\mathcal{C}=\mathcal{G}\times\mathcal{Z}$ is bounded and agent-visible: $g_k$ chooses the tool family, while $z_k$ grounds the selected tool family into a concrete scene-level subtask. This formalization turns long-horizon robot control into repeated bounded tool invocations, where the key methodological question is how to construct a VLA tool family $\mathcal{T}$ whose members can be reliably invoked, monitored, and composed by the high-level agent.

\begin{figure*}[t]
    \centering
    \includegraphics[width=\linewidth]{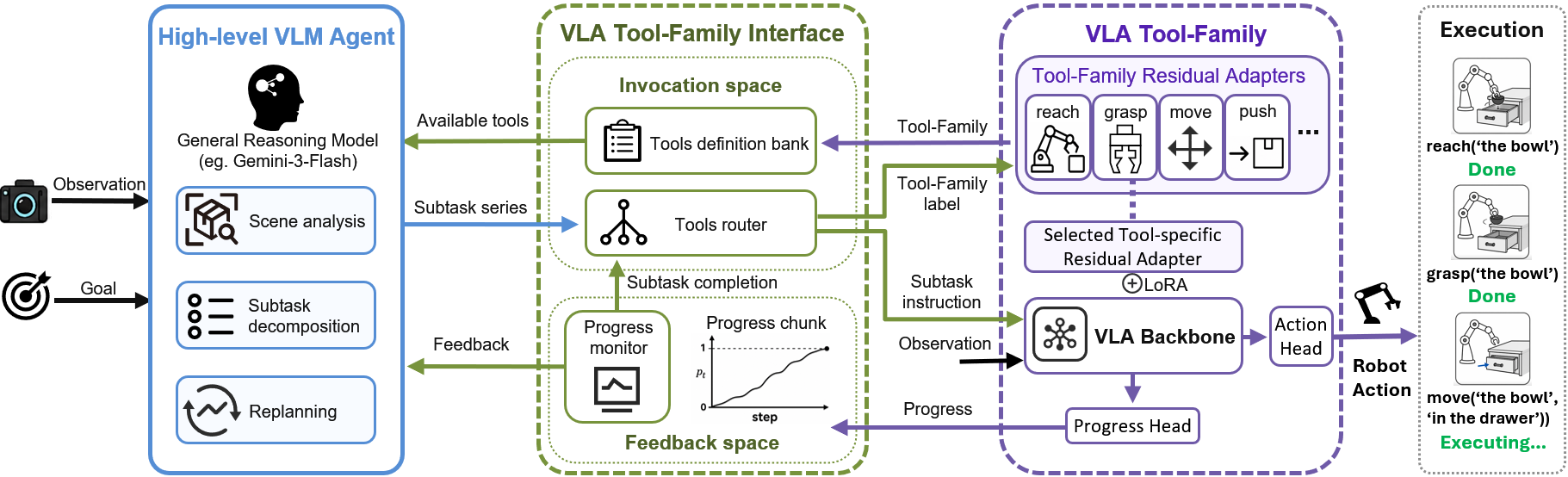}
    \caption{Overview of the closed-loop system with the VLAs-as-Tools strategy. A VLM agent performs high-level, long-horizon planning, while the VLA tool-family interface guides explicit tool calls and provides feedback for low-frequency agent intervention. VLAs become callable tools that faithfully execute invocations, with a parameter-efficient tool family generated via residual adapters.}
\vspace{-5mm}
    \label{fig:method-main}
\end{figure*}

\vspace{-2mm}
\section{Methods}
\vspace{-2mm}
\label{sec:method-main}

We instantiate the formal strategy above with two concrete design choices. First, we make a VLA callable through a VLA tool-family interface: the agent selects a tool family, provides a scene-grounded local instruction, and invokes the VLA for a bounded execution window. The VLA tool is therefore not a standalone task policy, but a bounded executor that receives an agent-specified invocation and returns progress feedback. Second, we train the VLA to follow this interface through Tool-Aligned Post-Training (TAPT), which aligns the training data, adapter structure, and optimization objectives with the same invocation unit used at test time.

\vspace{-2mm}
\subsection{Bidirectional Agent--Tool Interaction via a VLA Tool-Family Interface}
\vspace{-2mm}
\label{sec:interface}

Section~\ref{sec:vla-as-tool-formalization} formulates the agent--tool interface as
\(\mathcal{I}=(\mathcal{C},\mathcal{R})\). We instantiate this interface
for VLAs-as-tools through two typed message sets:
\(\mathcal{C}\) contains invocation messages sent by the high-level agent to a selected VLA
tool, and \(\mathcal{R}\) contains feedback messages returned by the tool for monitoring.
\vspace{-2mm}
\subsubsection{Invocation Messages $\mathcal{C}$}
\vspace{-2mm}
Each VLA tool call has two fields: a tool-family label and a scene-grounded instruction. The tool-family label \(g_k\in\mathcal{G}\), such as \textit{reach}, \textit{grasp}, or \textit{pull}, selects one callable member from the VLA tool family \(\mathcal{T}=\{T_g\}_{g\in\mathcal{G}}\). The taxonomy \(\mathcal{G}\) is data-driven: following primitive-level formulations~\citep{rh20tp}, it separates free-space motion from contact interaction, yielding 11 families on DROID, 7 of which are used in each of LIBERO and RoboTwin (Appendix~\ref{app:taxonomy}). The instruction \(z_k\in\mathcal{Z}\) then grounds the selected tool family in the current scene by specifying the object, relation, and desired local effect. Together, these two fields form the invocation message \(c_k=(g_k,z_k)\), so \(\mathcal{C}=\mathcal{G}\times\mathcal{Z}\).

This factorization makes the agent's request inspectable. The tool-family label \(g_k\) selects the tool family, while \(z_k\) grounds that tool family in the current scene. This is useful because visually similar instructions may require different tool families: \textit{grasp}, \textit{pull}, and \textit{push} can involve overlapping objects but differ in action distributions and termination conditions. The explicit tool-family label also gives TAPT a stable signal for organizing data and parameters, around tool-family specialization.

\vspace{-2mm}
\subsubsection{Feedback Messages $\mathcal{R}$}
\vspace{-2mm}

Feedback messages specify what the selected VLA tool reports while it is executing an invocation. We use a continuous progress signal \(p_t\in[0,1]\) as the main feedback message, where values near \(0\) indicate little progress and values near \(1\) indicate that the current invocation is nearly complete. This signal is local to the bounded call \(c_k=(g_k,z_k)\): it measures progress on the selected tool invocation, not success of the full long-horizon task.
Progress feedback is useful because the agent needs timely information before a subtask fully succeeds or fails. Repeatedly querying a large VLM to inspect execution is costly, while a binary completion signal can delay recovery. In contrast, a tool-side progress signal exposes intermediate execution states, such as stagnation or deviation, without requiring the agent to reason over every low-level action. In the interface, temporal progress is stored in the progress chunk, while a threshold-based monitor determines whether to advance, replan, or continue the current invocation.
The VLA predicts progress with an auxiliary head \(\hat{p}_t=\psi_{\omega}(b_t)\) attached to the backbone feature \(b_t\), with supervision described in Section~\ref{sec:tapt} and implementation details in Appendix~\ref{app:progress-details}.

\vspace{-2mm}
\subsection{Improving VLA Tool Usability with Tool-Aligned Post-Training}
\vspace{-2mm}
\label{sec:tapt}

Building on the VLA tool-family interface in Section~\ref{sec:interface}, this section describes how we train a base VLA to behave like a reliable callable tool. TAPT follows the same pipeline as the agent--tool loop. First, it rewrites both imitation and reinforcement-learning supervision around bounded invocations: demonstrations are segmented into invocation-labeled action windows, and RL rollouts are initialized and rewarded at the same subtask level. Second, it makes the tool-family label executable by routing each invocation through tool-family residual adapters on top of a shared VLA backbone. Third, it post-trains the resulting model to both execute the requested local behavior and predict progress feedback for that invocation. Standard downstream SFT and RL are then used as optimization settings for adapting and evaluating the same invocation-aligned tool family.

\vspace{-2mm}
\subsubsection{Invocation-Aligned Training Units}
\vspace{-2mm}
TAPT first rewrites imitation learning around the unit the agent calls at inference time. Instead of training only on full-task trajectories, we segment demonstrations into bounded windows and annotate each window with an invocation \(c=(g,z)\), where \(g\in\mathcal{G}\) identifies the tool family and \(z\in\mathcal{Z}\) describes the local scene effect. Each imitation unit is \(x_{\mathrm{IL}}=(o_{1:T}, z, g, a_{1:T}, p_{1:T}^{\star})\), where \(o_{1:T}\) and \(a_{1:T}\) are the observations and actions inside the window, and \(p_{1:T}^{\star}\) is a progress target or proxy. Segments are obtained from task annotations, state changes, contact events, and automatic multimodal labeling; the labeler assigns \(g\) using operational definitions based on contact pattern, force direction, and object motion. Detailed in Appendix~\ref{app:auto-labeling}.

The same invocation view is used for reinforcement learning. For each call \((g,z)\), we build a bounded subtask rollout rather than starting from the original full-task initial state. The rollout starts from states where that invocation should begin, obtained from demonstration boundaries or successful executions of preceding subtasks. It is evaluated with a local completion predicate \(\psi_{z,g}(s)\), such as contact, containment, relative pose, joint displacement, or object-on-surface relations. The rollout receives reward \(1\) if this predicate becomes true within horizon \(H\), and \(0\) otherwise; the episode ends when the predicate is satisfied or the horizon is reached.

Thus IL and RL share the same target: reliable execution of the current bounded invocation. IL teaches the actions and progress for a call, while RL tests whether the same call can be completed from states where the agent would invoke it.

\vspace{-2mm}
\subsubsection{Tool-Family Residual Parameterization}
\vspace{-2mm}

The tool-family label \(g\) should control the VLA's execution path, not merely appear as an extra language token. Tool families such as grasping, releasing, pulling, and rotating may share the same scene context, but they require different action distributions, termination conditions, and progress semantics. We therefore implement the VLA tool family with a shared pretrained backbone and deterministic tool-family residual adapters selected by \(g\).

Concretely, let \(f_{\Theta}\) be a pretrained VLA backbone. For each selected linear layer \(W\in\mathbb{R}^{d_{\mathrm{out}}\times d_{\mathrm{in}}}\), we add a bank of low-rank residual adapters \(\{\Delta W_g\}_{g\in\mathcal{G}}\), with one adapter per tool family:
\begin{equation}
    \Delta W_g = B_g A_g,
    \qquad
    \mathrm{rank}(\Delta W_g)\leq r\ll\min(d_{\mathrm{in}},d_{\mathrm{out}}).
\end{equation}
Given hidden activation \(x\) and selected family \(g\), the adapted layer computes \(y = W x + \Delta W_g x\). Thus \(g\) deterministically routes the call to a lightweight LoRA residual~\citep{hu2021lora}.

At inference time, the agent provides \(c_k=(g_k,z_k)\). The selected residual path remains active for the duration of the bounded invocation, and actions are produced as
\begin{equation}
    a_t \sim \pi_{\theta,\phi_{g_k}}(\cdot \mid o_t, z_k, h_t),
\end{equation}
where \(\phi_{g_k}\) denotes the residual-adapter parameters for family \(g_k\). The same backbone features are also used by the progress head defined in Section~\ref{sec:interface}. The adapters give each tool family a distinct execution path while preserving shared visual-language representations. Because each adapter is low-rank, adding tool families introduces only a small parameter overhead. A new family can also be added by training only its adapter while freezing the backbone and prior adapters, which retains previously learned skills far better than end-to-end sequential fine-tuning (Appendix~\ref{app:tool-expansion}).

\vspace{-2mm}
\subsubsection{Tool-Aligned Post-Training Objective}
\vspace{-2mm}

TAPT post-trains the parameterization above on a broad invocation-aligned corpus \(\mathcal{D}_{\mathrm{mid}}^{\mathrm{inv}}\) before benchmark-specific adaptation. The goal is to establish reusable semantics for the interface call \(c=(g,z)\): \(g\) selects an execution path, \(z\) grounds that path in the scene, and the progress target teaches the feedback signal returned to the agent.

For imitation learning, TAPT combines two supervised signals. The action-cloning term trains the selected residual path \(\phi_g\) to reproduce the actions for the invocation, while the progress term trains the feedback head to predict invocation progress:
\begin{equation}
    \mathcal{L}_{\mathrm{TAPT}}
    =
    \mathbb{E}_{(o,z,g,a,p^{\star})\sim\mathcal{D}_{\mathrm{mid}}^{\mathrm{inv}}}
    \left[
    -\sum_{t=1}^{T}\log\pi_{\theta,\phi_g}(a_t\mid o_t,z,h_t)
    +
    \frac{\lambda_{\mathrm{prog}}}{T}
    \sum_{t=1}^{T}\left\|\hat{p}_t-p_t^{\star}\right\|_2^2
    \right].
\end{equation}

\vspace{-2mm}
\begin{table*}[t]
\centering
\caption{Main results under imitation learning. We report success rate (SR) on each benchmark; red values in parentheses indicate the absolute improvement over the corresponding single-model SFT baseline with the same VLA backbone. TAPT denotes Tool-Aligned Post-Training.}
\vspace{-2mm}
\label{tab:main-il}
\small
\modernTableSetup
\setlength{\tabcolsep}{4pt}
\begin{tabularx}{\textwidth}{l l L
  S[table-format=2.1,table-number-alignment=right,table-column-width=0.1\textwidth] >{\raggedright\arraybackslash}p{0.085\textwidth}
  S[table-format=2.2,table-alignment-mode=marker,table-column-width=0.1\textwidth] >{\raggedright\arraybackslash}p{0.085\textwidth}}
\toprule
\rowcolor{TableHeader}
\textbf{Base Model} & \textbf{Adaptation} & \textbf{Deployment mode}
& \multicolumn{2}{c}{\textbf{LIBERO-Long SR} $\uparrow$}
& \multicolumn{2}{c}{\textbf{RoboTwin-SA SR} $\uparrow$} \\
\midrule
& SFT  & Standalone                & 77.2 & & 1.88  & \\
& SFT  & Direct planner            & 77.0 & & 13.80 & \\
\rowcolor{TableHighlight}
\multirow{-3}{*}{OpenVLA}
& TAPT & VLA tool-family interface & 82.4 & \gain{5.2} & 5.00 & \gain{3.12} \\
\midrule
& SFT  & Standalone                & 92.0 & & 16.88 & \\
& SFT  & Direct planner            & 80.2 & & 18.80 & \\
\rowcolor{TableHighlight}
\multirow{-3}{*}{OpenVLA-OFT}
& TAPT & VLA tool-family interface & 95.6 & \gain{3.6} & 48.75 & \gain{31.87} \\
\midrule
& SFT  & Standalone                & 92.4 & & 39.38 & \\
& SFT  & Direct planner            & 80.4 & & 30.00 & \\
\rowcolor{TableHighlight}
\multirow{-3}{*}{$\pi_{0.5}$}
& TAPT & VLA tool-family interface & 97.2 & \gain{4.8} & 62.50 & \gain{23.12} \\
\bottomrule
\end{tabularx}
\vspace{-3mm}
\end{table*}

\begin{table*}[t]
\centering
\caption{Results on modern long-horizon manipulation benchmarks. RoboMemArena is evaluated on 26 tasks $\times$ 10 seeds, and RoboLab-120 on 120 tasks $\times$ 5 seeds. ``Partial progress'' measures fine-grained task progress based on subtask completion.}
\label{tab:modern-long-horizon}
\small
\modernTableSetup
\setlength{\tabcolsep}{4pt}
\begin{tabularx}{\textwidth}{L l
  S[table-format=2.2,table-number-alignment=right,table-column-width=0.1\textwidth] >{\raggedright\arraybackslash}p{0.085\textwidth}
  S[table-format=2.2,table-number-alignment=right,table-column-width=0.1\textwidth] >{\raggedright\arraybackslash}p{0.085\textwidth}}
\toprule
\rowcolor{TableHeader}
\textbf{Benchmark} & \textbf{Deployment}
& \multicolumn{2}{c}{\textbf{Full-task SR} $\uparrow$}
& \multicolumn{2}{c}{\textbf{Partial progress} $\uparrow$} \\
\midrule
& Standard $\pi_{0.5}$ & 26.15 & & 48.99 & \\
\rowcolor{TableHighlight}
\multirow{-2}{*}{RoboMemArena~\citep{robomemarena}}
& \textbf{VLAs-as-Tools} & \bfseries 30.00 & \gain{3.85} & \bfseries 58.16 & \gain{9.17} \\
\midrule
& Standard $\pi_{0.5}$ & 28.5 & & 36.7 & \\
\rowcolor{TableHighlight}
\multirow{-2}{*}{RoboLab-120~\citep{robolab}}
& \textbf{VLAs-as-Tools} & \bfseries 34.8 & \gain{6.3} & \bfseries 42.8 & \gain{6.1} \\
\bottomrule
\end{tabularx}
\vspace{-3mm}
\end{table*}
\vspace{-2mm}
This objective jointly updates the tool-family residual adapters and progress head, so the resulting VLA learns both sides of the interface: executing the invocation and returning progress feedback.

For reinforcement learning, TAPT keeps the same invocation unit and optimizes the bounded subtask reward defined above. The reward evaluates only whether the current invocation is completed, rather than whether the full long-horizon task succeeds. In experiments we instantiate this RL stage with GRPO~\citep{shao2024deepseekmath}; implementation details are provided in Appendix~\ref{app:tapt-details}. Downstream SFT and RL reuse the same interface and objectives, replacing \(\mathcal{D}_{\mathrm{mid}}^{\mathrm{inv}}\) with benchmark-specific invocation-aligned data or environments. The result is a VLA tool family \(\mathcal{T}_{\theta,\Phi}=\{T_g\}_{g\in\mathcal{G}}\) that accepts \(c=(g,z)\), executes the selected tool, and returns progress feedback.

\vspace{-2mm}
\section{Experiment}
\vspace{-2mm}

\subsection{Experimental Setup}

Our experiments ask whether VLAs-as-tools improves long-horizon performance in simulation and the real world, whether TAPT makes VLA execution more faithful to agent invocations, and which components drive the gains; Section~\ref{sec:appendix-results} points to further analyses in the appendix.

We evaluate our method mainly with two representative VLA backbones, OpenVLA-OFT and $\pi_{0.5}$, and include OpenVLA as an additional supervised-adaptation baseline when useful~\citep{openvla,openvlaoft,pi05}. Experiments are conducted on LIBERO~\citep{libero}, RoboTwin~\citep{robotwin}, and CALVIN~\citep{calvin}, with additional long-horizon evaluation on RoboMemArena~\citep{robomemarena} and RoboLab-120~\citep{robolab}. For LIBERO, we use LIBERO-Long to highlight performance on long-horizon tasks. Since our data-splitting pipeline targets single-arm manipulation, RoboTwin evaluation uses 8 single-arm-executable tasks with a Franka embodiment, denoted RoboTwin-SA; the task list and selection rule are provided in Appendix~\ref{app:eval-protocol}. For CALVIN, we use CALVIN\_D with an 80\%/20\% train-test episode split, treating it as a complementary testbed with existing subtask-level structure. TAPT includes two invocation-aligned (IA) stages: \emph{IA Post-train}, an intermediate post-training stage on DROID-split, and \emph{IA SFT}, downstream supervised adaptation on benchmark-specific datasets such as LIBERO-Long-split and RoboTwin-split. The ``-split'' suffix denotes demonstrations processed by our data-splitting pipeline into invocation-aligned subtasks. In the IL training of both OpenVLA-OFT and $\pi_{0.5}$, IA Post-train is conducted for 1 epoch; during IA SFT, LIBERO is trained for 150K/30K steps and RoboTwin for 60K/30K steps, respectively, with batch sizes of 8/256 following the official configurations. All RL experiments are implemented with RLinf, with implementation details provided in Appendix~\ref{app:tapt-details}. Success rate denotes task-level success  with evaluation protocol details provided in Appendix~\ref{app:eval-protocol}.

\vspace{-2mm}
\subsection{Can the VLAs-as-tools Strategy Improve Long-Horizon Embodied Performance?}
\vspace{-2mm}

\begin{figure*}[t]
\centering
\begin{minipage}[t]{0.65\textwidth}
\centering
\textbf{(a) LIBERO-Long}

\vspace{0.5em}
\small
\modernTableSetup
\setlength{\tabcolsep}{2.5pt}
\begin{tabularx}{\linewidth}{l L S[table-format=2.1]}
\toprule
\rowcolor{TableHeader}
\textbf{Base model} & \textbf{Adaptation type} & {\textbf{Success Rate} $\uparrow$} \\
\midrule
            & Standard RL & 78.8 \\
            & Invocation-aligned RL & 79.0 \\
            & TAPT & 80.0 \\
\rowcolor{TableHighlight}
\multirow{-4}{*}{OpenVLA-OFT}
            & TAPT + VLA tool-family interface & \bfseries 82.6 \\
\midrule
            & Standard RL & 80.0 \\
            & Invocation-aligned RL & 89.0 \\
            & TAPT & 89.2 \\
\rowcolor{TableHighlight}
\multirow{-4}{*}{$\pi_{0.5}$}
            & TAPT + VLA tool-family interface & \bfseries 91.2 \\
\bottomrule
\end{tabularx}
\end{minipage}
\hfill
\begin{minipage}[t]{0.34\textwidth}
\centering
\textbf{(b) CALVIN}

\vspace{0.5em}
\includegraphics[width=\linewidth,height=3.6cm,keepaspectratio]{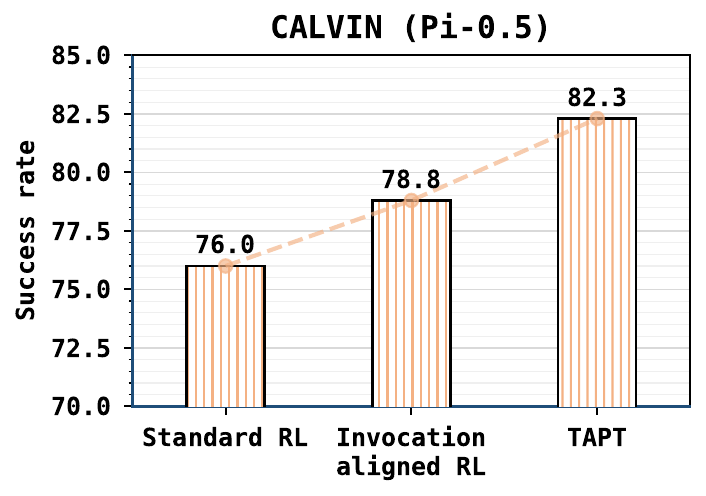}
\end{minipage}
\vspace{-3mm}
\caption{Main reinforcement-learning results on LIBERO-Long and CALVIN. LIBERO-Long evaluates standard RL, TAPT, and VLA tool-family interface deployment; CALVIN tests whether the gains persist under the benchmark's native subtask structure rather than our split construction.}
\vspace{-3mm}
\label{fig:main-rl}
\end{figure*}
\begin{table*}[t]
\centering
\caption{Invocation fidelity on LIBERO-IF benchmark under TAPT component ablations. Faithful Rate (\%) measures completion of the invoked counterfactual goal, while Non-biased Rate (\%) measures avoidance of the original LIBERO-10 source-task bias.}
\vspace{-2mm}
\label{tab:invocation-fidelity}
\small
\modernTableSetup
\setlength{\tabcolsep}{3pt}
\begin{tabularx}{\textwidth}{c c >{\centering\arraybackslash}X
  *{4}{S[table-format=2.1,table-column-width=0.13\textwidth]}}
\toprule
\rowcolor{TableHeader}
\multicolumn{3}{c}{\textbf{TAPT components}}
& \multicolumn{2}{c}{\textbf{OpenVLA-OFT}}
& \multicolumn{2}{c}{\textbf{\boldmath$\pi_{0.5}$}} \\
\cmidrule(lr){1-3}\cmidrule(lr){4-5}\cmidrule(lr){6-7}
\rowcolor{TableHeader}
IA SFT & IA Post-train & Residual adapters
& {Faithful $\uparrow$} & {Non-biased $\uparrow$}
& {Faithful $\uparrow$} & {Non-biased $\uparrow$} \\
\midrule
 &  &  & 19.4 & 31.2 & 24.8 & 39.6 \\
\checkmark &  &  & 30.4 & 38.8 & 36.6 & 43.2 \\
\checkmark & \checkmark &  & 40.6 & 39.2 & 38.2 & 46.6 \\
\rowcolor{TableHighlight}
\checkmark & \checkmark & \checkmark & \bfseries 54.0 & \bfseries 47.4 & \bfseries 54.8 & \bfseries 54.6 \\
\bottomrule
\end{tabularx}
\vspace{-5mm}
\end{table*}

\begin{table*}[t]
\centering
\caption{Core ablation of TAPT and the VLA tool-family interface on LIBERO-Long with OpenVLA-OFT and \(\pi_{0.5}\); TAPT includes IA SFT/Post-train and tool-family residual adapters.}
\vspace{-2mm}
\label{tab:core-ablation}
\small
\modernTableSetup
\setlength{\tabcolsep}{3pt}
\begin{tabularx}{\textwidth}{c c >{\centering\arraybackslash}X >{\centering\arraybackslash}X
  *{2}{S[table-format=2.1,table-column-width=0.15\textwidth]}}
\toprule
\rowcolor{TableHeader}
\multicolumn{4}{c}{\textbf{Components}} & \multicolumn{2}{c}{\textbf{Success Rate (\%)} $\uparrow$} \\
\cmidrule(lr){1-4}\cmidrule(lr){5-6}
\rowcolor{TableHeader}
IA SFT & IA Post-train & Tool-family residual adapters & VLA tool-family interface
& {OpenVLA-OFT} & {\(\pi_{0.5}\)} \\
\midrule
 &  &  &  & 92.0 & 92.4 \\
\checkmark &  &  &  & 93.0 & 93.4 \\
\checkmark & \checkmark &  &  & 94.2 & 91.2 \\
\checkmark & \checkmark & \checkmark &  & 94.8 & 96.0 \\
 &  &  & \checkmark & 80.2 & 80.4 \\
\rowcolor{TableHighlight}
\checkmark & \checkmark & \checkmark & \checkmark & \bfseries 95.6 & \bfseries 97.2 \\
\bottomrule
\end{tabularx}
\end{table*}

\paragraph{Supervised downstream adaptation.}
We first evaluate imitation-learning adaptation when downstream demonstrations are available. Table~\ref{tab:main-il} compares three deployment modes: a standalone SFT-trained VLA, a standard SFT-trained VLA with a direct planner, and our TAPT-trained VLA with the VLA tool-family interface. Deployment mode denotes how the trained VLA is used at evaluation time: \emph{Standalone} executes the VLA as a single task policy, \emph{Direct planner} lets a high-level planner periodically issue language instructions, and the \emph{VLA tool-family interface} establishes a tightly coupled bidirectional interaction between the agent planner and VLA tools. The results show that simply placing a standard SFT policy inside an agent loop is insufficient: on LIBERO-Long, direct planner deployment changes performance by $-0.2$, $-11.8$, and $-12.0$ points for OpenVLA, OpenVLA-OFT, and $\pi_{0.5}$, respectively. In contrast, TAPT with the VLA tool-family interface consistently improves over the corresponding standalone SFT baselines, with gains of up to $+5.2$ points on LIBERO-Long and $+31.87$ points on RoboTwin-SA. These results indicate that the main benefit does not come from wrapping a VLA with a planner alone, but from making the VLA executor tool-aligned and reliable under agent--tool interaction.

\vspace{-1mm}
\paragraph{Reinforcement learning in the target domain.}
We next evaluate whether \emph{VLAs-as-Tools} remains effective when VLAs are improved through target-domain RL. Figure~\ref{fig:main-rl}(a) compares Standard RL, IA RL, TAPT, and TAPT with the VLA tool-family interface on LIBERO-Long. Here, Standard RL denotes full-task RL that optimizes the original long-horizon task success without IA units. The full configuration outperforms Standard RL by $+3.8$ and $+11.2$ points for OpenVLA-OFT and $\pi_{0.5}$, respectively. The VLA tool-family interface further adds $+2.6$ and $+2.0$ points over TAPT alone, showing the benefit of high-level planning and agent--tool interaction.
Figure~\ref{fig:main-rl}(b) evaluates CALVIN under its native subtask structure, where success also increases from $76.0$ to $78.8$ with Invocation-aligned RL and further to $82.3$ with full TAPT. These results show that the proposed strategy also benefits RL-based training, and that the gains are not tied to our own split construction.

\vspace{-2mm}
\subsection{Do the Gains Extend to Modern Long-Horizon Benchmarks?}
\label{sec:modern-long-horizon}
\vspace{-2mm}

RoboMemArena tests memory-dependent multi-stage execution, while RoboLab-120 tests visual, relational, and procedural generalization~\citep{robomemarena,robolab}. Under a uniform 2,500-step budget, Table~\ref{tab:modern-long-horizon} shows that \emph{VLAs-as-Tools} improves RoboMemArena full-task SR by 3.85 points and partial progress by 9.17 points, and improves RoboLab-120 by 6.3 and 6.1 points, respectively. In a representative RoboMemArena recovery, nominal tool calls terminate before the physical goal is satisfied; the agent detects the mismatch, regenerates the remaining suffix, and re-invokes the failed pick-and-place to complete the task.

\vspace{-2mm}
\subsection{Can VLAs-as-Tools Improve Real-World Long-Horizon Manipulation?}
\label{sec:real-world}
\vspace{-2mm}

\begin{wraptable}{r}{0.47\textwidth}
\vspace{-3mm}
\centering
\caption{Real-world successes and success rates (\%) over 20 trials per task.}
\label{tab:real-world}
\small
\modernTableSetup
\setlength{\tabcolsep}{3pt}
\begin{tabularx}{\linewidth}{L r r
  >{\columncolor{TableHighlight}}r >{\columncolor{TableHighlight}}r}
\toprule
\rowcolor{TableHeader}
\textbf{Task} & \multicolumn{2}{c}{\textbf{Fine-tuned $\pi_{0.5}$}} & \multicolumn{2}{c}{\textbf{Ours}} \\
\midrule
Simple      & 4/20  & (20.0) & \textbf{13/20} & \textbf{(65.0)} \\
Conditioned & 1/20  & (5.0)  & \textbf{13/20} & \textbf{(65.0)} \\
Long        & 5/20  & (25.0) & \textbf{9/20}  & \textbf{(45.0)} \\
\midrule
Avg.        & 10/60 & (16.7) & \textbf{35/60} & \textbf{(58.3)} \\
\bottomrule
\end{tabularx}
\vspace{-4mm}
\end{wraptable}

\begin{figure*}[t]
\centering
\includegraphics[width=\textwidth]{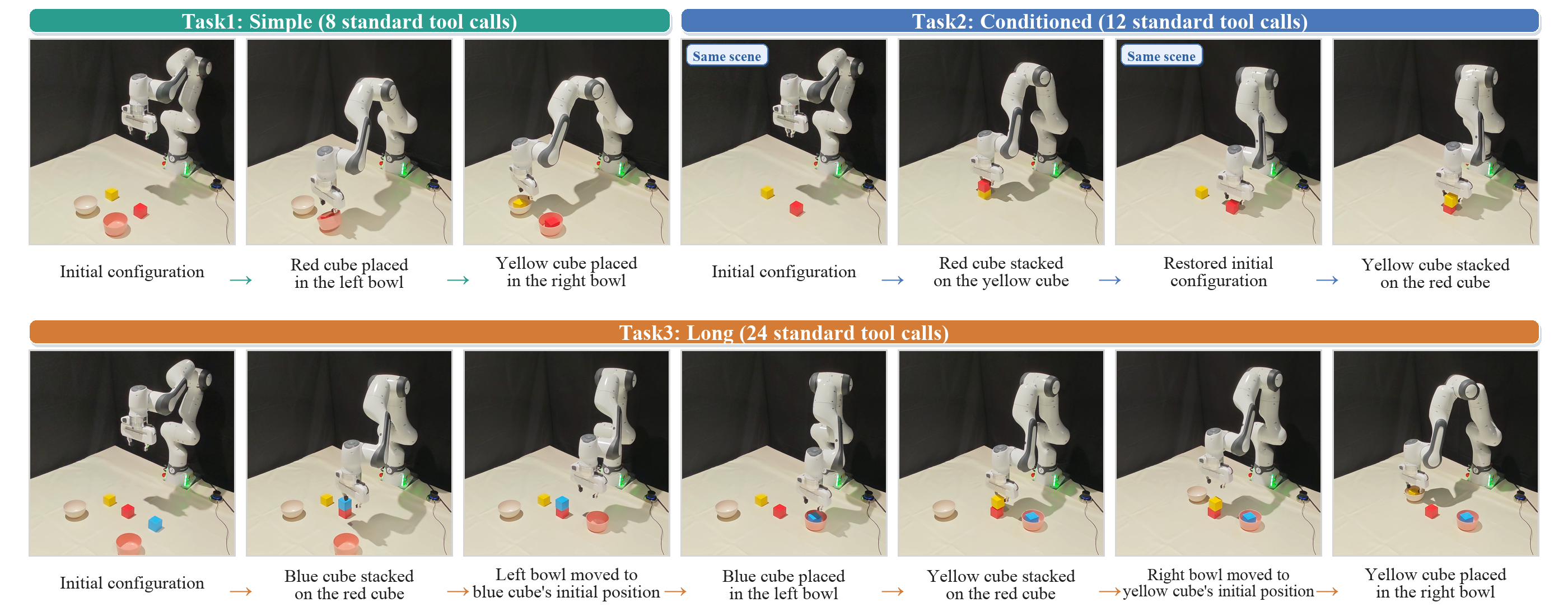}
\vspace{-5mm}
\caption{Stage-by-stage real-world task progression. Top: Task1 (Simple) and Task2 (Conditioned); bottom: Task3 (Long). Each square crop shows a completed physical stage; Task2 marks its repeated physical configuration as ``Same scene.''}
\label{fig:real-world-rollouts}
\vspace{-3mm}
\end{figure*}

We compare a fine-tuned $\pi_{0.5}$ baseline with \emph{VLAs-as-Tools} on a Franka Research 3 using one fixed front-view RealSense D435. Both use 30 demonstrations per task, 30K training steps on eight A100s, one RTX 4090 at inference, and the same 20 marked initial configurations. Success is judged by the final physical state (Appendix~\ref{app:real-world}). We evaluate three real-world tasks---\emph{Simple}, \emph{Conditioned}, and \emph{Long}---as illustrated in Figure~\ref{fig:real-world-rollouts}.

Table~\ref{tab:real-world} shows that average success rises from 16.7\% to 58.3\%. Conditioned provides the clearest evidence, improving from 5.0\% to 65.0\%: the same visual input can occur at different task stages that require different next actions. Lacking explicit task state, the end-to-end policy cannot reliably disambiguate these aliased states or choose the action that advances execution. Our planner resolves this ambiguity by tracking completed stages and invoking the appropriate tool. Appendix Figure~\ref{fig:real-conditioned-stage} visualizes this stage-tracking failure.

\vspace{-2mm}
\subsection{Does TAPT Make VLA Execution More Faithful to Agent Invocations?}
\vspace{-2mm}

Long-horizon success alone does not show whether a VLA has become a more reliable tool for an agent. We thus evaluate invocation fidelity on LIBERO-Instruction Following (LIBERO-IF), a counterfactual suite built on familiar LIBERO-10 scenes. LIBERO-IF follows LIBERO-CF~\citep{LibroCF} but focuses on our agent--tool setting: each task changes the invoked goal---for example, the target object, spatial relation, or stopping point---with the high-level plan fixed, varying only the executor by progressively enabling TAPT components. Detailed settings are provided in Appendix~\ref{app:libero-if}.

Faithful Rate measures completion of the invoked counterfactual goal. Biased Rate measures whether execution instead pursues the original source-task goal, and Non-biased Rate is $100\%-\text{Biased Rate}$. Table~\ref{tab:invocation-fidelity} shows that full TAPT improves Faithful/Non-biased Rate by $+34.6/+16.2$ points for OpenVLA-OFT and $+30.0/+15.0$ points for \(\pi_{0.5}\). Thus, TAPT not only raises success but also strengthens the invocation--behavior binding needed for reliable tool use: the executor follows the requested call more often and is less likely to revert to its training-task sequence.
\vspace{-3mm}
\subsection{Core Design Ablation}
\vspace{-2mm}
\label{sec:ablation}

Table~\ref{tab:core-ablation} separates the effects of executor alignment and agent--tool interaction. Invocation-aligned SFT adds 1.0 point for both backbones. Further post-training improves OpenVLA-OFT to 94.2\%, but reduces \(\pi_{0.5}\) from 93.4\% to 91.2\%, likely because the open-source post-training data is less matched to the high-quality training distribution of \(\pi_{0.5}\). Tool-family residual adapters then raise OpenVLA-OFT to 94.8\% and recover the backbone-specific drop in \(\pi_{0.5}\), raising its success rate to 96.0\%. The interface alone reaches only 80.2\%/80.4\%, showing that planner wrapping cannot make an unaligned executor reliable; combining TAPT with the interface instead achieves the best 95.6\%/97.2\%. These results confirm that invocation-aligned execution and bidirectional agent--tool interaction are complementary.

\begin{figure*}[t]
    \centering
    \includegraphics[width=\textwidth]{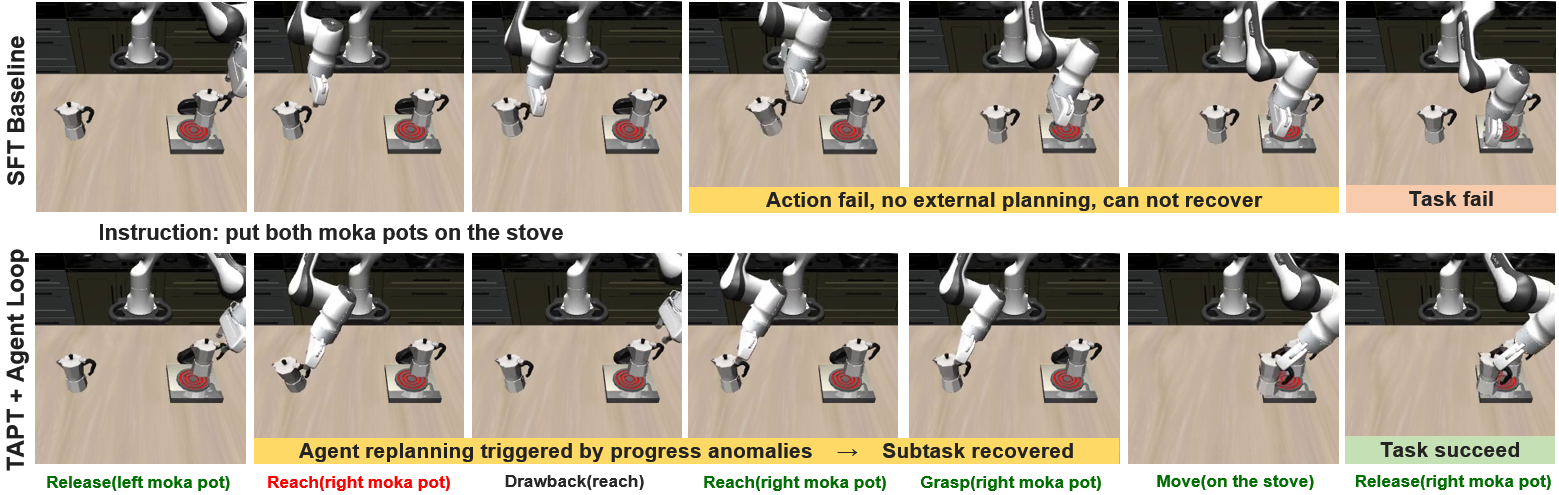}
    \vspace{-4mm}
    \caption{
    Qualitative comparison of long-horizon failure recovery. Our method prevents cascading failures through progress feedback and replanning, turning local errors into recoverable transitions.
    }
    \label{fig:task8_replan_recovery}
    \vspace{-6mm}
\end{figure*}

\vspace{-3mm}
\subsection{Qualitative Analysis}
\vspace{-2mm}

Figure~\ref{fig:task8_replan_recovery} illustrates how the agent--tool loop prevents a local execution error from cascading through a long-horizon task. After dropping the left moka pot, the baseline continues the nominal trajectory from an invalid state and ultimately fails, exposing the inability of a monolithic rollout to revise later actions after an early state change. In contrast, our agent detects the failed grasp through progress feedback, withdraws from the failed interaction, and re-invokes the corresponding VLA tool before proceeding to the next pot. The system thus resolves the error as a bounded local correction rather than requiring the executor to recover the entire task within one uninterrupted rollout.

\vspace{-3mm}
\subsection{Additional Results in the Appendix}
\label{sec:appendix-results}
\vspace{-2mm}

The appendix further reports \textbf{few-shot adaptation} (up to $+11.5$ points; Table~\ref{tab:fewshot}), \textbf{planner robustness} across four VLMs (Table~\ref{tab:brain-ablation}), \textbf{planning efficiency} of progress-triggered replanning (Table~\ref{tab:libero_replan_ablation}), \textbf{tool-family extensibility} (Appendix~\ref{app:taxonomy}), and \textbf{adapter cost} (only $+9\%$ parameters; Appendix~\ref{app:adapter-cost}).
\vspace{-3mm}
\section{Conclusion}
\vspace{-2mm}
We presented \textbf{VLAs-as-Tools}, a framework that treats VLA models as callable, specialized executors under a planning-capable multimodal agent. By separating high-level decomposition, progress monitoring, and recovery from bounded robot-control execution, the framework makes VLA policies more composable for long-horizon embodied tasks. We further introduced Tool-Aligned Post-Training to improve invocation fidelity and specialize VLA tools without discarding their pretrained semantic capabilities. Experiments in simulation and on a real Franka robot show gains in task success, instruction alignment, few-shot adaptation, and planner robustness. The real-world study covers one robot, one camera, and three tabletop tasks, and primitives absent from pretraining and adaptation data still require new demonstrations; broader settings remain future work.

\bibliography{references}
\bibliographystyle{references}

\clearpage
\appendix
\section*{Appendix}

\section{Progress Supervision and Feedback}
\label{app:progress-details}

\paragraph{Supervision and prediction.}
For each bounded subtask invocation, we supervise a scalar progress signal using a normalized temporal proxy. For an automatically segmented execution window of length \(T\), the frame-level target is \(p_t^{\star}=(t-1)/(T-1)\), increasing linearly from \(0\) at the first frame to \(1\) at the last. The progress head pools the final-layer backbone feature and passes it through a lightweight \(D\!\rightarrow\!1024\!\rightarrow\!256\!\rightarrow\!1\) MLP followed by a sigmoid. It is jointly optimized with the action objective using mean-squared error and \(\lambda_{\mathrm{prog}}=0.1\). At runtime, each 2\,Hz VLA inference jointly predicts an action chunk and its aligned progress feedback, requiring no separate perception pass or model call.

\begin{figure*}[t]
    \centering
    \includegraphics[width=\textwidth]{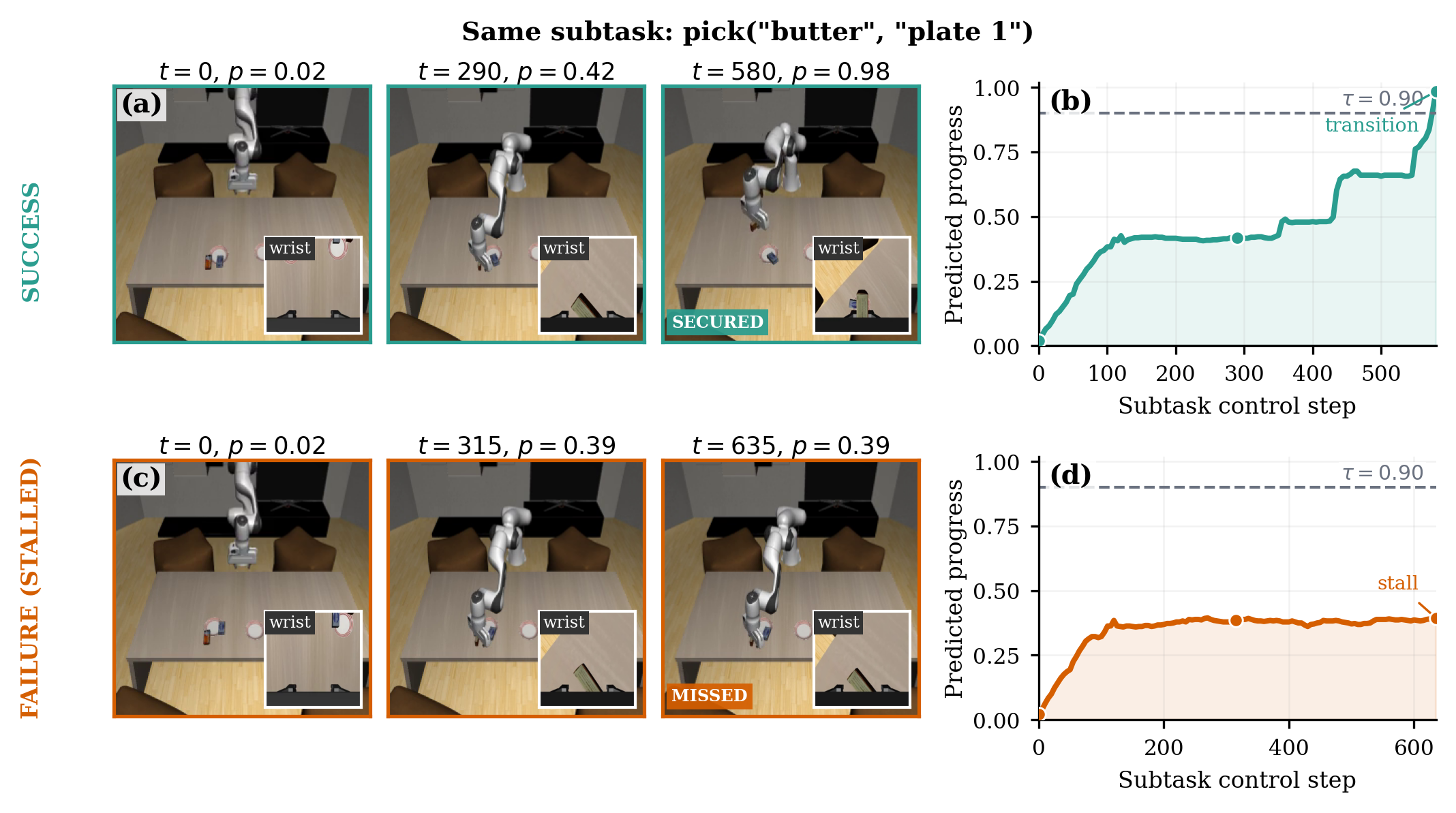}
    \caption{Local progress feedback for the same bounded pick invocation. In the successful rollout (top), predicted progress rises as the grasp develops and crosses the completion threshold \(\tau=0.90\), triggering a transition to the next tool. In the failed rollout (bottom), the missed grasp causes progress to plateau, allowing the runtime monitor to identify stagnation and request replanning.}
    \label{fig:progress-feedback}
\end{figure*}

\paragraph{Runtime semantics and control.}
Progress is local to the currently active subtask invocation rather than a global task-completion percentage. The VLM planner instead maintains global execution state from the plan, current observations, completed calls, and execution history. This separation avoids requiring the low-level executor to infer an ambiguous global completion state while exposing intervention points before failure or timeout. A runtime monitor combines a completion threshold with consecutive-sample smoothing and lookback-based progress gains to decide whether to continue the current call, advance to the next tool, or invoke replanning after stagnation. Figure~\ref{fig:progress-feedback} illustrates these outcomes for the same pick invocation: progress crosses the completion threshold after a successful grasp but plateaus when the grasp is missed. Because supervision comes from temporal position within segmented windows, the prediction is a learned temporal completion proxy correlated with physical execution state, not a perfectly calibrated estimate of semantic task completion.

\clearpage
\section{Automatic Labeling Pipeline}
\label{app:auto-labeling}

\begin{figure*}[t]
    \centering
    \includegraphics[width=\textwidth]{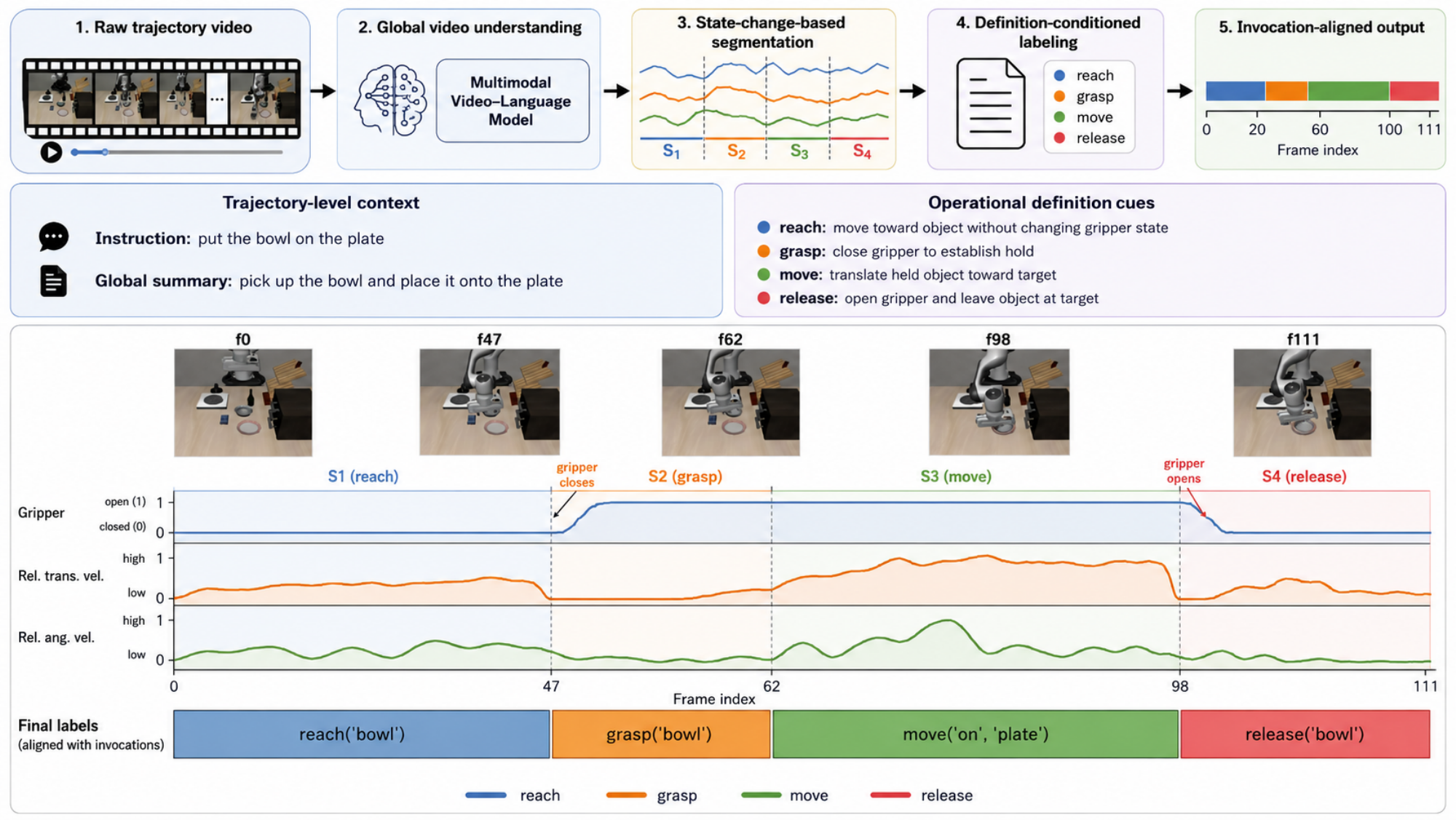}
    \caption{
    Visualization of the automatic labeling pipeline output. Raw robot trajectories are segmented into bounded manipulation clips and assigned invocation-aligned tool-family labels with normalized progress annotations, providing structured examples for Tool-Aligned Post-Training and downstream split adaptation.
    }
    \label{fig:visualize_data_pipeline}
\end{figure*}

We convert raw robot trajectories into invocation-aligned training examples in three stages: global video understanding, state-change-based segmentation, and definition-conditioned tool-family labeling. Unlike trajectory-level captioning used in large robot datasets~\citep{walke2023bridgedata,khazatsky2024droid,openx2024}, our pipeline assigns each bounded segment an operational label defined by its contact pattern, motion, physical effect, and termination condition. Figure~\ref{fig:visualize_data_pipeline} summarizes the resulting clips, labels, and normalized progress annotations.

\paragraph{Global video understanding.}
For each trajectory, a multimodal model summarizes the manipulated objects, task intent, spatial relations, and major contact events. This global context disambiguates short local clips during segmentation and labeling.

\paragraph{State-change-based segmentation.}
We split trajectories at gripper open--close transitions and motion changes. Gripper transitions provide explicit boundaries for grasp and release, while relative translational and angular velocities between the end-effector and object provide supporting cues for push, pull, flip, rotate, insert, press, and strike. These cues propose boundaries but do not determine labels by themselves.

\paragraph{Definition-conditioned tool-family labeling.}
For each candidate segment, the model receives the local video, global summary, state-change cues, and the operational definitions in Table~\ref{tab:tool-family-definitions}. It selects the family that best matches the observed contact, motion direction, object displacement, and termination condition.

\begin{table}[t]
\centering
\caption{Operational definitions used by the automatic labeling pipeline. Arguments indicate the manipulated object, target object or target position, and geometric parameter when applicable.}
\label{tab:tool-family-definitions}
\small
\modernTableSetup
\renewcommand{\arraystretch}{1.25}
\setlength{\tabcolsep}{4pt}
\begin{tabularx}{\textwidth}{l L}
\toprule
\rowcolor{TableHeader}
\textbf{Tool family} & \textbf{Definition} \\
\midrule
\texttt{reach(obj)} & The arm moves toward the object while the gripper state remains unchanged. The segment ends before grasping, contact-driven translation, or another object-changing interaction begins. \\
\texttt{grasp(obj)} & The gripper closes on the object after the end-effector is positioned around it, establishing a stable hold. The segment ends when the gripper closure completes and the object moves with the end-effector. \\
\texttt{move(pos, obj2)} & The arm translates toward a target position while the gripper state remains unchanged. Here \texttt{obj2} denotes the reference object or target area and \texttt{pos} denotes the target position relative to it. \\
\texttt{release(obj)} & The gripper opens to detach the held object at its current location. The segment ends when the gripper is open and the object no longer moves with the end-effector. \\
\texttt{flip(obj, angle)} & The robot rotates the object, typically around a horizontal axis, so that its vertical orientation is inverted or the opposite side is exposed. \\
\texttt{rotate(obj, angle)} & The robot adjusts the object's 3D pose or orientation without vertical inversion, or manipulates an articulated mechanism through a rotational joint. \\
\texttt{push(obj, pos)} & The end-effector contacts the object and translates while applying forward force, causing the object to move toward the target position. The action ends when object motion stops and contact is released. \\
\texttt{pull(obj, pos)} & The end-effector establishes a pulling contact and translates backward, causing the object to move toward the target position. \\
\texttt{insert(obj1, obj2)} & \texttt{obj1} is aligned with an opening, slot, or receptacle of \texttt{obj2}, then translated along the insertion axis until it is partially or fully seated. \\
\texttt{press(obj)} & The end-effector moves downward to contact the object and applies sustained normal force, usually with limited tangential motion. \\
\texttt{strike(obj1, obj2, pos)} & \texttt{obj1}, the end-effector, or another held item is moved rapidly to impact \texttt{obj2} or a specified location \texttt{pos}. \\
\bottomrule
\end{tabularx}
\end{table}

\paragraph{Signal use and consistency checks.}
Gripper transitions determine candidate interaction boundaries; motion cues then distinguish families with similar gripper states. If local evidence conflicts with the global context or a family definition, the pipeline lowers confidence or sends the segment for manual review rather than forcing a label. The final examples are $x_{\mathrm{IL}}=(o_{1:T}, z, g, a_{1:T}, p_{1:T}^{\star})$, where $z$ is the local invocation instruction, $g$ is the tool-family label, and $p_{1:T}^{\star}$ is normalized within-segment progress.

\section{Tool-Family Taxonomy and Extensibility}
\label{app:taxonomy}

Our tool-family taxonomy is data-driven. Following prior primitive-level formulations~\citep{rh20tp}, we distinguish free-space motion primitives from interaction primitives that involve physical contact. Applying the labeling pipeline in Appendix~\ref{app:auto-labeling} to broad-coverage datasets such as DROID yields 11 fundamental tool families (Table~\ref{tab:tool-family-definitions}), 7 of which are used in each of LIBERO and RoboTwin.

\subsection{Taxonomy Granularity}
\label{app:taxonomy-granularity}

To test whether performance depends on the exact granularity of the taxonomy, we compare the current tool families with a coarser, task-level taxonomy on LIBERO-Long using \(\pi_{0.5}\) (Table~\ref{tab:taxonomy-granularity}). The modest gap indicates that performance is not overly sensitive to the exact taxonomy: our data-derived tool families work without benchmark-specific or exhaustive manual design, while separating motion and contact primitives provides a useful inductive bias.

\begin{table}[t]
\centering
\caption{Effect of tool-family granularity on LIBERO-Long with \(\pi_{0.5}\).}
\label{tab:taxonomy-granularity}
\small
\modernTableSetup
\setlength{\tabcolsep}{4pt}
\begin{tabularx}{0.75\textwidth}{l L S[table-format=2.1,table-column-width=0.2\textwidth]}
\toprule
\rowcolor{TableHeader}
\textbf{Taxonomy} & \textbf{Tool families} & {\textbf{LIBERO-Long SR} $\uparrow$} \\
\midrule
Coarse & pick, place, open, close, rotate & 96.6 \\
\rowcolor{TableHighlight}
Current & reach, grasp, move, release, pull, push, rotate & \bfseries 97.2 \\
\bottomrule
\end{tabularx}
\end{table}

\subsection{Incremental Tool-Family Expansion}
\label{app:tool-expansion}

We distinguish recombinations of existing primitives from genuinely new physical primitives. If a task can be expressed as a new sequence of registered families, the high-level planner can solve it by composing existing tools with scene-grounded instructions. However, if it requires a primitive absent from both the pretrained VLA and the adaptation data, no data-driven VLA, whether end-to-end or tool-based, can be expected to execute it reliably zero-shot. This is a general coverage limit of data-driven robot policies~\citep{qwenrobotmanip}, not of our taxonomy.

VLAs-as-Tools becomes advantageous once demonstrations of the new primitive become available. Instead of updating the full policy, we add a lightweight family-specific adapter while freezing the shared backbone and all prior adapters. This isolates the update, reduces interference with existing families, and avoids data replay. We test this property in a three-stage LIBERO-Long experiment:
\begin{enumerate}
    \item \textbf{Stage 1 (initial registry):} learn reach, grasp, move, release, and rotate.
    \item \textbf{Stage 2 (add pull):} add opening tasks that need a new pull primitive; train only its adapter.
    \item \textbf{Stage 3 (add push):} add closing tasks that need a new push primitive; train only its adapter.
\end{enumerate}
At Stages 2 and 3, the backbone and prior adapters stay frozen, with no data replay. We compare this modular strategy with end-to-end sequential fine-tuning, which updates the shared policy whenever a new primitive is added. We report five metrics: \emph{S2 New}, the success of the new pull family after Stage 2; \emph{S2 Ret.}, the retention of the five Stage 1 families; \emph{S3 New}, the success of the new push family after Stage 3; \emph{S3 Ret.}, the retention of all previously learned families; and \emph{Final}, the LIBERO-Long success rate with all seven families.

\begin{table}[t]
\centering
\caption{Incremental tool-family expansion on LIBERO-Long. New families are added in two stages without data replay.}
\label{tab:tool-expansion}
\small
\modernTableSetup
\setlength{\tabcolsep}{4pt}
\begin{tabularx}{\textwidth}{L *{5}{S[table-format=2.1,table-column-width=0.09\textwidth]}}
\toprule
\rowcolor{TableHeader}
\textbf{Method} & {\textbf{S2 New}} & {\textbf{S2 Ret.}} & {\textbf{S3 New}} & {\textbf{S3 Ret.}} & {\textbf{Final}} \\
\midrule
End-to-end sequential FT & \bfseries 98.6 & 91.4 & \bfseries 98.8 & 86.4 & 82.4 \\
\rowcolor{TableHighlight}
Family-specific LoRA expansion (ours) & 98.4 & \bfseries 99.0 & \bfseries 98.8 & \bfseries 97.6 & \bfseries 95.0 \\
\bottomrule
\end{tabularx}
\end{table}

Table~\ref{tab:tool-expansion} shows that both methods learn pull and push, confirming that new operations require training data. The key difference is prior-skill retention: after Stage 3, end-to-end fine-tuning retains 86.4\% of prior skills, while ours retains 97.6\% and achieves 95.0\% final LIBERO-Long success, consistent with prior adapter-based continual learning~\citep{tail,clare}.

\section{Additional Details for Tool-Aligned Post-Training}
\label{app:tapt-details}

This appendix records the formal details that are omitted from the main Method section for readability. For reinforcement learning, each invocation \((z,g)\) defines a bounded subtask MDP whose reset distribution starts from states where that invocation should begin. Let \(\mathcal{S}^{0}_{z,g}\) be the pool of such states, obtained from demonstration boundaries or successful executions of preceding subtasks. We sample
\begin{equation}
    \rho_{z,g}(s)=
    \frac{1}{|\mathcal{S}^{0}_{z,g}|}
    \sum_{\bar{s}\in\mathcal{S}^{0}_{z,g}}\delta(s=\bar{s}),
    \qquad
    s_0\sim\rho_{z,g}.
\end{equation}
Each invocation has a state-based completion predicate \(\psi_{z,g}(s)\in\{0,1\}\). For a bounded rollout \(\tau=(s_0,a_0,\ldots,s_H)\), the invocation-level reward is
\begin{equation}
    R(\tau;z,g)=
    \mathbf{1}\!\left[\max_{0\leq t\leq H}\psi_{z,g}(s_t)=1\right].
\end{equation}
Thus the RL stage optimizes the expected completion of the current invocation,
\begin{equation}
    \max_{\theta,\Phi}
    \mathbb{E}_{(z,g),\,\tau\sim\pi_{\theta,\phi_g}(\cdot\mid z),\,\mathcal{M}_{z,g}}
    \left[
    R(\tau;z,g)
    \right],
\end{equation}
which we instantiate with GRPO in our experiments.

\paragraph{Completion predicates in RLinf.}
We instantiate \(\psi_{z,g}\) using simulator-state predicates. 
For spatial relations that are already part of the LIBERO benchmark definition, such as placing an object \texttt{in} a receptacle or \texttt{on} a target surface, we reuse LIBERO's native success predicates. 
For intermediate invocations that are not exposed as standalone LIBERO task goals, such as reaching or grasping, we add lightweight predicates over robot and object state. 
This keeps the final task relations aligned with LIBERO while still allowing RL to optimize bounded intermediate tool calls.

\begin{table}[t]
\centering
\small
\caption{Completion predicates used to instantiate \(\psi_{z,g}\) in RLinf. We reuse LIBERO native predicates for benchmark-defined spatial relations and add state-based predicates for intermediate tool invocations.}
\modernTableSetup
\renewcommand{\arraystretch}{1.25}
\setlength{\tabcolsep}{4pt}
\begin{tabularx}{\textwidth}{l l L}
\toprule
\rowcolor{TableHeader}
\textbf{Invocation family \(g\)} & \textbf{Predicate source} & \textbf{Success condition for \(\psi_{z,g}\)} \\
\midrule
Reach & Added state predicate & The end-effector is within a task-specific distance threshold of the target object, handle, drawer, or knob. \\
Grasp & Added state predicate & The target object is held by the gripper, verified using contact and gripper-state information. \\
Move & LIBERO native predicate & The target object satisfies LIBERO's native containment predicate with the target receptacle or on-top predicate with the target surface. \\
Release & Added state predicate & The gripper is open and the object is no longer held; for placement subtasks, the required spatial relation must remain satisfied. \\
Rotate & Added state predicate & The relevant articulated state, such as a stove-knob angle, crosses the task-specific threshold. \\
Push & Added state predicate & The object is displaced into the target region or satisfies the corresponding scene-state condition. \\
\bottomrule
\end{tabularx}
\label{tab:rlinf-predicates}
\end{table}

\paragraph{Reset-state pool refresh.}
The choice of \(\mathcal{S}^{0}_{z,g}\) is important for composing independently trained subtask policies. 
In the first round of subtask RL, we collect reset states using the initial policy or demonstration boundaries. 
This provides valid states for training individual subtasks, but it can become stale after policy improvement: once a preceding subtask policy is updated, it may end with different gripper poses, object contacts, approach directions, or small object displacements than those in the original pool. 
Consequently, a downstream subtask may be trained on a reset distribution that no longer matches the states produced by the learned composed policy. 
In our experiments, this mismatch can make individually improved subtasks compose worse, sometimes reducing full-task success.

To reduce this distribution shift, we refresh the reset-state pools after one round of subtask RL. 
We deploy the trained tool-family subtask policy, save boundary states whenever the preceding invocation succeeds, and use these refreshed states as \(\mathcal{S}^{0}_{z,g}\) for the next round of subtask training. 
The refreshed pool therefore matches the state distribution induced by the current composed policy, improving the handoff between neighboring invocations. 
Empirically, this refresh step recovers full-task performance gains that are not obtained from stale initial-policy pools alone.

\begin{table}[t]
\centering
\small
\caption{Reset-state pool choices in RLinf. Refreshing \(\mathcal{S}^{0}_{z,g}\) with the trained tool-family policy reduces the distribution shift introduced when subtask RL changes the states passed between invocations.}
\modernTableSetup
\renewcommand{\arraystretch}{1.25}
\setlength{\tabcolsep}{4pt}
\begin{tabularx}{\textwidth}{>{\raggedright\arraybackslash\hsize=0.8\hsize}X
  >{\raggedright\arraybackslash\hsize=1.05\hsize}X
  >{\raggedright\arraybackslash\hsize=1.15\hsize}X}
\toprule
\rowcolor{TableHeader}
\textbf{Reset-state pool \(\mathcal{S}^{0}_{z,g}\)} & \textbf{What it matches} & \textbf{Observed effect} \\
\midrule
Original task resets & The benchmark initial state distribution & Appropriate for the first invocation, but unsuitable for later subtasks because the policy must implicitly reproduce the task prefix. \\
Initial-policy or demonstration boundary pool & States where each subtask should begin before subtask RL changes the policy & Enables first-round subtask RL, but can become mismatched after upstream subtasks are updated. \\
\rowcolor{TableHighlight}
Refreshed tool-family boundary pool & Boundary states produced by the trained composed subtask policy & Improves subtask handoff and increases full-task success by aligning training resets with deployment states. \\
\bottomrule
\end{tabularx}
\label{tab:rlinf-state-pool-refresh}
\end{table}

For the tool-family residual parameterization, each selected linear layer \(W\in\mathbb{R}^{d_{\mathrm{out}}\times d_{\mathrm{in}}}\) receives a family-specific LoRA residual \(\Delta W_g=B_gA_g\), where \(A_g\in\mathbb{R}^{r\times d_{\mathrm{in}}}\), \(B_g\in\mathbb{R}^{d_{\mathrm{out}}\times r}\), and \(r\ll\min(d_{\mathrm{in}},d_{\mathrm{out}})\). 
Given hidden activation \(x\), the adapted layer computes \(y=Wx+\Delta W_gx\). 
For selected layers \(\mathcal{L}\) and \(K=|\mathcal{G}|\) tool families, this adds \(\mathcal{O}(|\mathcal{L}|K r(d_{\mathrm{in}}+d_{\mathrm{out}}))\) parameters while keeping the pretrained VLA backbone shared across families.

\section{Additional Adaptation and Planner Robustness Results}
\label{app:adaptation-planner-results}

\subsection{Tool-Family Adapter Cost}
\label{app:adapter-cost}

Table~\ref{tab:normalized-cost} reports the normalized parameter and inference cost of tool-family specialization, which is lightweight. Adapters add only 9\% parameters and 7\% inference time to OpenVLA-OFT; adapted \(\pi_{0.5}\) remains at 0.50$\times$ the OpenVLA-OFT parameter baseline despite higher latency, avoiding a separate full VLA per tool family.

\begin{center}
\centering
\captionof{table}{Normalized model size and inference time on LIBERO-Long. Values are normalized to OpenVLA-OFT without tool-family adapters.}
\label{tab:normalized-cost}
\vspace{2.5mm}
\small
\modernTableSetup
\setlength{\tabcolsep}{4pt}
\begin{tabularx}{0.7\textwidth}{L *{2}{>{\centering\arraybackslash}p{0.17\textwidth}}}
\toprule
\rowcolor{TableHeader}
\textbf{Model} & \textbf{Parameters} & \textbf{Inference time} \\
\midrule
OpenVLA-OFT & 1.00\(\times\) & 1.00\(\times\) \\
OpenVLA-OFT + tool families & 1.09\(\times\) & 1.07\(\times\) \\
\(\pi_{0.5}\) & 0.44\(\times\) & 1.26\(\times\) \\
\(\pi_{0.5}\) + tool families & \textbf{0.50\(\times\)} & 1.34\(\times\) \\
\bottomrule
\end{tabularx}
\end{center}

A central purpose of Tool-Aligned Post-Training is to make the VLA interface easier to adapt to a downstream benchmark. We therefore evaluate few-shot invocation-aligned SFT on LIBERO-Long under limited downstream demonstrations. If the invocation-aligned post-training stage has established reusable invocation semantics, the model should require substantially fewer downstream demonstrations to recover strong downstream performance. Table~\ref{tab:fewshot} shows that TAPT improves downstream sample efficiency, with the largest gain in the most data-limited OpenVLA-OFT 10-shot setting ($+11.5$ points) and consistent gains for \(\pi_{0.5}\) under both 10-shot and 20-shot adaptation ($+6.4$ and $+5.6$ points).

\begin{table}[t]
\centering
\caption{Few-shot downstream adaptation on LIBERO-Long. Results report success rate under limited SFT demonstrations.}
\label{tab:fewshot}
\small
\modernTableSetup
\setlength{\tabcolsep}{4pt}
\begin{tabularx}{0.7\textwidth}{l L *{2}{S[table-format=2.1,table-column-width=0.13\textwidth]}}
\toprule
\rowcolor{TableHeader}
\textbf{Base model} & \textbf{Method} & {\textbf{10-shot}} & {\textbf{20-shot}} \\
\midrule
            & Baseline & 8.0 & 33.8 \\
\rowcolor{TableHighlight}
\multirow{-2}{*}{OpenVLA-OFT}
            & VLAs-as-Tools & 19.5 & 36.4 \\
\midrule
            & Baseline & 43.4 & 60.0 \\
\rowcolor{TableHighlight}
\multirow{-2}{*}{\(\pi_{0.5}\)}
            & VLAs-as-Tools & 49.8 & 65.6 \\
\bottomrule
\end{tabularx}
\end{table}

We additionally test whether the agent--tool interface depends on a particular large VLM planner. In this experiment, the OpenVLA-OFT VLA tool families are fixed and only the high-level VLM is changed. Table~\ref{tab:brain-ablation} shows that the interface is not tied to a single high-level planner: across four strong VLM planners, LIBERO-Long success varies by only $0.6$ points.

\begin{table}[t]
\centering
\caption{High-level VLM planner ablation on LIBERO-Long under the OpenVLA-OFT VLA tool families.}
\label{tab:brain-ablation}
\small
\modernTableSetup
\setlength{\tabcolsep}{4pt}
\begin{tabularx}{0.7\textwidth}{l L S[table-format=2.1,table-column-width=0.2\textwidth]}
\toprule
\rowcolor{TableHeader}
\textbf{Model type} & \textbf{Large VLM} & {\textbf{Success Rate} $\uparrow$} \\
\midrule
\multirow{2}{*}{Closed-source} & Gemini 3 Flash & 95.6 \\
& GPT 5.4 & 95.4 \\
\midrule
\multirow{2}{*}{Open-source} & Qwen3.5-397B-A17B & 95.4 \\
& Qwen3VL-235B-A22B & 95.0 \\
\bottomrule
\end{tabularx}
\end{table}

\subsection{Agent Planning Efficiency}
\label{app:planning-efficiency}

Table~\ref{tab:libero_replan_ablation} compares progress-triggered replanning with direct VLM monitoring every five steps. The call counts measure VLM requests during execution and exclude initial plan construction: the tool-family interface counts replanning requests, while direct monitoring counts periodic VLM queries. For OpenVLA, OpenVLA-OFT, and \(\pi_{0.5}\), the average runtime calls decrease from $109.5/58.2/57.2$ to $1.988/0.304/0.228$ per episode. Intervention frequency is the fraction of episodes with at least one replanning request; replan success rate is the fraction of these intervened episodes that ultimately succeed. The corresponding replan success rates of $47.1\%/34.4\%/61.4\%$ show that structured progress feedback supports recovery with few runtime VLM queries.

\begin{table}[t]
\centering
\caption{
Ablation of agent-level replanning on LIBERO-Long. Calls are averaged over evaluation episodes and count runtime replanning or monitoring requests; initial plan construction is excluded.
}
\label{tab:libero_replan_ablation}
\small
\modernTableSetup
\setlength{\tabcolsep}{4pt}
\begin{tabularx}{\textwidth}{L
  S[table-format=2.1,table-column-width=0.18\textwidth]
  S[table-format=2.1,table-column-width=0.15\textwidth]
  S[table-format=1.3,table-column-width=0.16\textwidth]
  S[table-format=3.1,table-column-width=0.235\textwidth]}
\toprule
\rowcolor{TableHeader}
& \multicolumn{3}{c}{\textbf{with VLA tool-family interface}}
& \multicolumn{1}{c}{\textbf{Direct VLM monitoring}} \\
\cmidrule(lr){2-4}\cmidrule(lr){5-5}
\rowcolor{TableHeader}
\multirow{-2}{*}{\textbf{Model}}
& {Intervene Freq. (\%)} & {Replan SR (\%)} & {\shortstack{Avg. Runtime\\Calls}} & {\shortstack{Avg. Runtime\\Calls}} \\
\midrule
OpenVLA     & 40.8 & 47.1 & 1.988 & 109.5 \\
OpenVLA-OFT & 12.2 & 34.4 & 0.304 & 58.2 \\
$\pi_{0.5}$ & 11.4 & 61.4 & 0.228 & 57.2 \\
\bottomrule
\end{tabularx}
\end{table}

\section{Real-World Evaluation Details}
\label{app:real-world}

\paragraph{Robot and sensing setup.}
We use a Franka Research 3 robot and a single Intel RealSense D435 RGB-D camera, with no wrist-mounted camera. The camera is placed opposite the robot and angled downward to cover an approximately $40\times40$~cm workspace on a white tabletop against a black background.

\paragraph{Training protocol.}
Human experts collect 30 demonstrations per task using a 3D mouse, producing three task-specific datasets. For each task, the fine-tuned $\pi_{0.5}$ baseline and \emph{VLAs-as-Tools} are trained separately on the same 30 demonstrations for 30K steps using eight A100 GPUs. Both methods run inference on a single NVIDIA RTX 4090. Thus, the comparison controls the real-robot data, training budget, and inference hardware.

\paragraph{Evaluation protocol.}
Each method is evaluated over 20 rollouts per task. We pre-sample and physically mark 20 randomized initial object configurations, and use the same configurations for both methods. Object positions vary within the camera view while preserving the approximate left-to-right ordering required by each task. The maximum execution time is 60 seconds for Simple and Conditioned and 120 seconds for Long. Success is determined using the task-specific completion criteria below.

\paragraph{Simple.}
Red and yellow cubes are placed in front of the robot, with two bowls farther away. The robot must place the red cube into the left bowl and then the yellow cube into the right bowl. A rollout succeeds only if both cubes remain stably inside their designated bowls. This task evaluates multi-stage decomposition and execution stability.

\paragraph{Conditioned.}
Red and yellow cubes are initially placed on the left and right. The robot must stack the left red cube on the right yellow cube, remove it to restore the initial layout, and then stack the right yellow cube on the left red cube. The scene therefore returns to a visually similar state while requiring a different subsequent action. A rollout succeeds only if both stacks are completed stably. This task evaluates instruction following and resistance to scene-conditioned visual shortcuts.

\paragraph{Long.}
Blue, red, and yellow cubes are placed from left to right, with two bowls farther away. The robot must (1) stack blue on red; (2) move the left bowl to the blue cube's original position; (3) place blue into that bowl; (4) stack yellow on red; (5) move the right bowl to the yellow cube's original position; and (6) place yellow into the right bowl. The task comprises 24 sequential tool invocations. A rollout succeeds only if both bowls reach their designated positions and the blue and yellow cubes remain stably inside them. This task evaluates extended planning, tool composition, and recovery.

\begin{center}
\centering
\includegraphics[width=\textwidth]{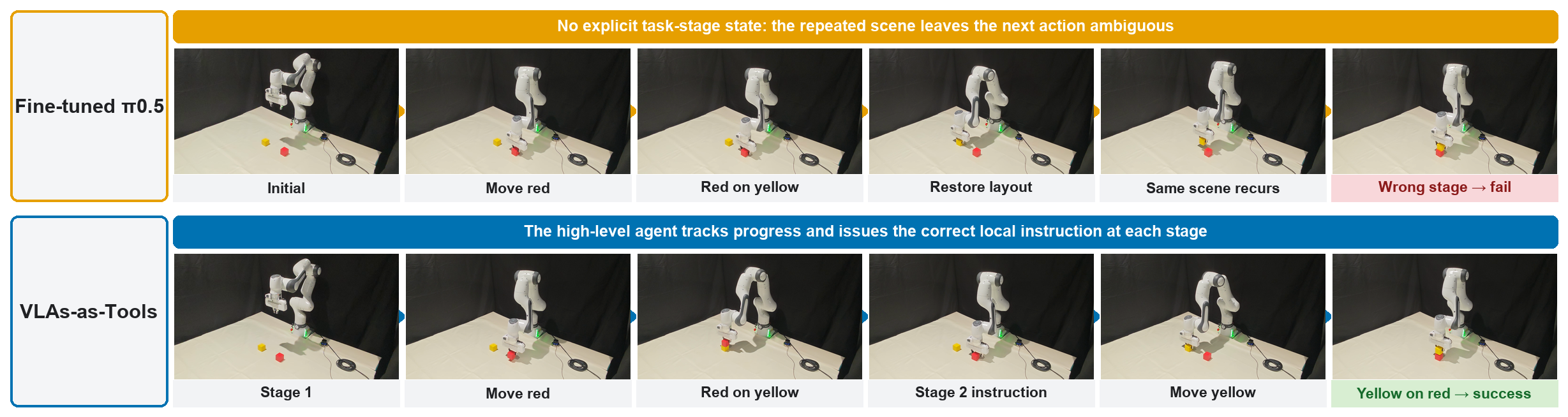}
\vspace{-3mm}
\captionof{figure}{Conditioned-task rollouts. After the layout is restored, the repeated observation leaves the fine-tuned VLA without an explicit task-stage cue. VLAs-as-Tools tracks progress in the high-level agent and issues the correct stage-specific VLA invocation.}
\label{fig:real-conditioned-stage}
\end{center}

Figure~\ref{fig:real-conditioned-stage} illustrates the largest real-world gain: the monolithic baseline repeats the wrong stage after the scene recurs, whereas the agent remembers the completed prefix and instructs the VLA to execute the opposite stack.

\section{Evaluation Protocol Details}
\label{app:eval-protocol}

This appendix summarizes the evaluation protocol used for the task-level success rates reported in the main experiments. Unless otherwise specified, all methods compared within the same benchmark use the same held-out tasks, initial-state files, rollout horizons, random seeds, and success predicates. Success rate is computed as the fraction of evaluation episodes that satisfy the benchmark-defined task predicate before timeout.
The trial counts for individual benchmarks are specified below.

\paragraph{LIBERO-Long.}
For task-level success evaluation, each policy is evaluated over 50 trials per task from the held-out LIBERO-Long initial states and is counted as successful if the native LIBERO task predicate is satisfied within the simulator's standard rollout horizon. The same task split and initial-state set are used for standalone VLA policies, direct-planner deployment, and VLA tool-family interface deployment.

\paragraph{RoboTwin-SA.}
Because our demonstration-splitting pipeline targets single-arm manipulation, we evaluate on RoboTwin-SA (Single-Arm), the subset of 8 RoboTwin tasks that can be executed with a single Franka arm. A task is included if its successful demonstrations do not require bimanual coordination and can be segmented into the tool-family invocation types used in our interface. The selected tasks are: 
\begin{quote}
\small
\texttt{adjust\_bottle}, 
\texttt{click\_alarmclock}, 
\texttt{move\_stapler\_pad}, 
\texttt{press\_stapler}, 
\texttt{beat\_block\_hammer}, 
\texttt{click\_bell}, 
\texttt{place\_empty\_cup}, 
\texttt{place\_shoe}.
\end{quote}

RoboTwin-SA evaluations are configured for 20 trials per task, or 160 planned rollouts per configuration. Success is measured by the benchmark task predicate at the end of each rollout or when the task terminates successfully.

\paragraph{CALVIN.}
For CALVIN, we use CALVIN\_D and split episodes into 80\% training and 20\% testing. The evaluation follows the benchmark's native subtask structure: a rollout is successful if it completes the required long-horizon instruction sequence under CALVIN's state-based success checks. In the RL experiments, the native subtask annotations are also used to define invocation-aligned resets and rewards, which lets us evaluate the proposed strategy without relying on our own split construction.

\paragraph{RoboMemArena.}
RoboMemArena~\citep{robomemarena} contains 26 long-horizon manipulation tasks that test temporal memory through multi-object counting, occlusion, ordered execution, and transferring; its trajectories average more than 1,000 environment steps. We evaluate all 26 simulated tasks with 10 seeded rollouts per task and a maximum horizon of 2,500 steps, using the same initial seeds and budget for the standard $\pi_{0.5}$ baseline and \emph{VLAs-as-Tools}. Full-task success requires every reference stage of an episode to be completed, while partial progress is the mean fraction of required stages completed before termination. We report both metrics averaged over all 260 episodes.

\paragraph{LIBERO-IF.}
\label{app:libero-if}
LIBERO-Instruction Following (LIBERO-IF) evaluates invocation fidelity rather than only task success. Drawing on the counterfactual evaluation of LIBERO-CF~\citep{LibroCF}, we build our own instruction-following benchmark tailored to our agent--tool setting, where the high-level plan is fixed and only the invoked VLA executor varies. Specifically, we construct 10 counterfactual tasks from LIBERO-10 layouts by modifying the requested goal while keeping the scene visually close to a familiar source task. We report Faithful Rate, the fraction of rollouts satisfying the active counterfactual goal, and Non-biased Rate, defined as \(100-\text{Biased Rate}\), where Biased Rate measures whether the rollout also satisfies the paired source LIBERO-10 goal during a post-success continuation window. The continuation window length is 100.

\paragraph{RoboLab-120.}
RoboLab-120~\citep{robolab} contains 120 manipulation tasks organized along three competency axes (visual, relational, and procedural) across three difficulty levels, testing generalization beyond the training distribution. All tasks use the benchmark's default language instructions. We evaluate all 120 tasks with 5 seeds per task and a maximum horizon of 2,500 steps, using the same initial seeds and budget for the standard $\pi_{0.5}$ baseline and \emph{VLAs-as-Tools}. Full-task success requires the benchmark task predicate to be satisfied before termination, while partial progress is the mean fraction of required subtasks completed. We report both metrics averaged over all 600 episodes.

\end{document}